\documentclass[conference]{IEEEtran}

\IEEEoverridecommandlockouts
\usepackage{cite}
\usepackage[colorlinks,urlcolor=blue,linkcolor=blue,anchorcolor=blue,citecolor=green]{hyperref}
\usepackage[table,x11names]{xcolor}
\usepackage{amsmath,amssymb,amsfonts}
\usepackage{algorithmic}
\usepackage{algorithm}
\usepackage{graphicx}
\usepackage{textcomp}
\usepackage{latexsym}
\usepackage{booktabs}
\usepackage{float}
\usepackage{placeins}
\usepackage{breqn}
\usepackage{makecell}
\usepackage{pifont}
\usepackage{multirow}
\usepackage{float}
\usepackage{colortbl}  
\usepackage{xcolor}
\usepackage{array}
\definecolor{maroon}{cmyk}{0,0.87,0.68,0.32}
\def\BibTeX{{\rm B\kern-.05em{\sc i\kern-.025em b}\kern-.08em
    T\kern-.1667em\lower.7ex\hbox{E}\kern-.125emX}}
    
\usepackage{color}
\IEEEoverridecommandlockouts
\usepackage{marvosym}
\usepackage{fancyhdr}

\makeatletter

\def\ps@IEEEtitlepagestyle{%
  \def\@oddfoot{\mycopyrightnotice}%
  \def\@evenfoot{}%
}
\def\mycopyrightnotice{%
  {\footnotesize 979-8-3503-3748-8/23/\$31.00~\copyright~2025 IEEE\hfill} 
  \gdef\mycopyrightnotice{}
}

\begin{document}

\title{A Hierarchical Geometry-guided Transformer for Histological Subtyping of Primary Liver Cancer}
\author{Paper ID: B700}

\DeclareRobustCommand*{\IEEEauthorrefmark}[1]{%
  \raisebox{0pt}[0pt][0pt]{\textsuperscript{\footnotesize #1}}%
}


\author{
    \IEEEauthorblockN{
        Anwen Lu\IEEEauthorrefmark{1},
        Mingxin Liu\IEEEauthorrefmark{1}, 
        Yiping Jiao\IEEEauthorrefmark{1},
        Hongyi Gong\IEEEauthorrefmark{1},
        Geyang Xu\IEEEauthorrefmark{3},
        Jun Chen\IEEEauthorrefmark{2}, and
        Jun Xu\IEEEauthorrefmark{1,\Letter}  \thanks{\textsuperscript{\Letter} Corresponding author: Jun Xu (jxu@nuist.edu.cn)} 
    }
    \IEEEauthorblockA{
        \IEEEauthorrefmark{1} Jiangsu Key Laboratory of Intelligent Medical Image Computing, School of Artificial Intelligence, \\ Nanjing University of Information Science and Technology, China \\
        \IEEEauthorrefmark{2} Department of Pathology, Nanjing Drum Tower Hospital, Affiliated Hospital of Medical School, Nanjing University, China \\
        \IEEEauthorrefmark{3} Department of Biostatistics, School of Public Health, 1415 Washington Heights, Ann Arbor, MI 48109-2029
    }
}

\maketitle

\thispagestyle{fancy}
\renewcommand{\headrulewidth}{0pt} 
\cfoot{}

\begin{abstract}
Primary liver malignancies are widely recognized as the most heterogeneous and prognostically diverse cancers of the digestive system. Among these, hepatocellular carcinoma (HCC) and intrahepatic cholangiocarcinoma (ICC) emerge as the two principal histological subtypes, demonstrating significantly greater complexity in tissue morphology and cellular architecture than other common tumors. The intricate representation of features in Whole Slide Images (WSIs) encompasses abundant crucial information for liver cancer histological subtyping, regarding hierarchical pyramid structure, tumor microenvironment (TME), and geometric representation. However, recent approaches have not adequately exploited these indispensable effective descriptors, resulting in a limited understanding of histological representation and suboptimal subtyping performance. To mitigate these limitations, \textbf{A} hie\textbf{R}archical \textbf{G}eometry-g\textbf{U}ided tran\textbf{S}former (ARGUS) is proposed to advance histological subtyping in liver cancer by capturing the macro-meso-micro hierarchical information within the TME. 
Specifically, we first construct a micro-geometry feature to represent fine-grained cell-level pattern via a geometric structure across nuclei, thereby providing a more refined and precise perspective for delineating pathological images.
Then, a Hierarchical Field-of-Views (FoVs) Alignment module is designed to model macro- and meso-level hierarchical interactions inherent in WSIs.
Finally, the augmented micro-geometry and FoVs features are fused into a joint representation via present Geometry Prior Guided Fusion strategy for modeling holistic phenotype interactions.
Extensive experiments on public and private cohorts demonstrate that our ARGUS achieves 
state-of-the-art (SOTA) performance in histological subtyping of liver cancer, which provide an effective diagnostic tool for primary liver malignancies in clinical practice. Related code will be available to public.
\end{abstract}


\section{Introduction}
Primary liver cancer is the fourth leading cause of cancer-related mortality worldwide and represents an increasingly critical public health concern~\cite{akinyemiju2017burden,bray2024global}. The two most prevalent subtypes are hepatocellular carcinoma (HCC), originating from the hepatocytes, and intrahepatic cholangiocarcinoma (ICC), arising from the biliary epithelial cells. These entities lie at opposite ends of the primary liver tumor spectrum, exhibiting distinct histopathological features and clinical behaviors~\cite{paradis2023pathogenesis}. Notably, ICC is an aggressive malignancy with highly heterogeneous, associated with poorer prognosis and greater histological complexity than HCC, thereby posing significant diagnostic challenges~\cite{hu2019comparative}. Accurate subtyping of ICC is therefore of considerable clinical importance, as it provides essential guidance for personalized treatment strategies.

In clinical practice, Alvaro et al.~\cite{european2023easl} demonstrated that ICC can be classified into three subtypes according to their biliary origin and histopathological features: large duct type, small duct type, and fine duct type. This classification framework-rooted in biliary developmental lineage, histological architecture, and molecular profiles, provides a more precise representation of the tumor’s biological behavior and clinical characteristics. Among these subtypes, the large duct type is typically associated with aggressive behavior and poor prognosis, whereas the fine duct type exhibits lower invasiveness and more favorable clinical outcomes. Consequently, accurate subtyping is crucial for informing treatment strategies and predicting patient prognosis. Recently, some studies provided primary liver cancer diagnostic solutions using radiology~\cite{calderaro2023deep} or histology images~\cite{song2025deep} for HCC vs. ICC or HCC fine-grained subtyping. However, few studies have investigated the subtyping of ICC using histopathological images, which exhibit high inter-subtype similarity and thus render this a challenging fine-grained classification task.

\label{sec:method}
\begin{figure*}[t]
    \centering
    \centerline{\includegraphics[width=1\textwidth]{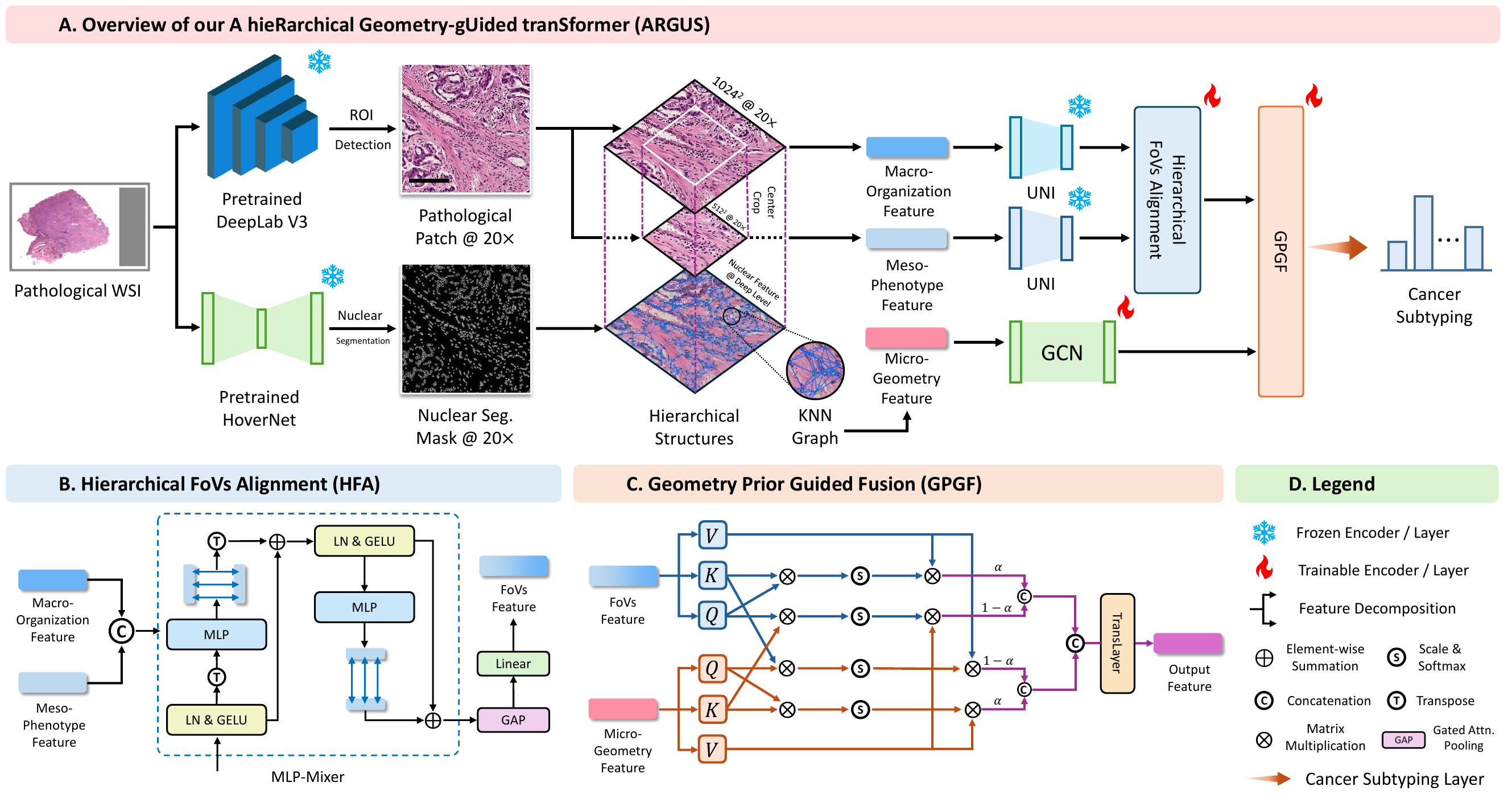}}
    \caption{Overview of the proposed ARGUS. (a) Overall workflow of our framework for histological subtyping, (b) Hierarchical FoVs Alignment (HFA) module, (c) Geometry Prior Guided Fusion (GPGF) module, (d) Legend for the symbols used.}
    \label{ARGUS}
\end{figure*}

Histological subtyping of primary liver cancer is a challenging task that requires focusing both coarse-grained features such as tumor size/invasion, lymphocytic infiltrates, and the broad organization of these phenotypes in the TME, also fine-grained morphological features such as nuclear atypia or tumor presence, for assessing precise subtyping of malignancy~\cite{chen2022scaling}. 
Recent bleeding-edge approaches in similar tasks almost adapt the multiple instance learning (MIL) framework~\cite{ilse2018attention,li2021dual,shao2021transmil,cai2024seqfrt,liu2023mgct,liu2025murrenet}, which are unable to capture important contextual and hierarchical information that have known great significance in cancer diagnosis~\cite{chen2021whole}. 
To this end, some studies proposed multi-scales/FoVs or graph-based models to tackle aforementioned issues. 
For example, Li et al.~\cite{li2021dual} designed dual-stream multiple instance learning (DSMIL) which leverages tissue features ranging from millimeter-scale to cellular-scale.
Chen et al.~\cite{chen2022scaling} introduced Hierarchical Image Pyramid Transformer (HIPT) to learn the hierarchical structure in WSIs using two levels of resolution in histopathological image representations via self-supervised learning.
While these methods are not context-aware and unable to model important morphological feature interactions between cell/nuclei identities and tissue types which are crucial for patient diagnosis~\cite{liu2024exploiting}. 
Therefore, many graph-based models were presented to leverage geometric features to represent the fine-grained cell-to-cell interactions under a higher resolution of WSI~\cite{chen2021whole,zheng2022graph,liu2024unleashing}.
Nevertheless, these graph-based models are usually using a coarse patch-based graph convolutional network (GCN) to extract the geometric representation in complicated histology WSIs, thus neglecting the rich information from shape, size, and other useful features of nuclei/cell identities~\cite{yang2025explainable}.

In this paper, we propose a graph-based, weakly-supervised framework, dubbed \textbf{A} hie\textbf{R}archical \textbf{G}eometry-g\textbf{U}ided tran\textbf{S}former (ARGUS), as shown in Fig.~\ref{ARGUS}, tailored for liver cancer histological subtyping by modeling hierarchical interactions across macro-meso-micro resolutions of pathological WSI. 

The \textbf{main contributions} of this paper are as follows: 
\begin{enumerate}
\item We represent the deepest FoV of WSIs via a micro geometric structure across nuclei identities using hand-crafted features, thereby providing a more fine-grained perspective for pathological image interpretation.
\item We introduce a Hierarchical FoVs Alignment module (HFA) as a multi-resolution feature aggregation approach to effectively capturing image representations of hierarchical structure in gigapixel WSIs.
\item We designed a Geometry Prior Guided Fusion strategy (GPGF) to integrate hierarchical morphological features and geometric representations to provide comprehensive learning of pathological images.
\item We performed extensive experiments on two datasets from The Cancer Genome Atlas (TCGA) and in-house collection, the results demonstrate that our method consistently outperforms current state-of-the-art methods.
\end{enumerate}

\section{Methodology}
\subsection{Data Preprocessing and Feature Extraction}
\subsubsection{Data Preprocessing}
We leveraged a pretrained tumor Region-of-Interest (ROI) segmentation model based on DeepLabv3~\cite{chen2017rethinking} to minimize the influence of benign and non-tissue regions. We processed the WSI using a sliding window approach, where each window was measured 562$\times$562 microns tissue area. These patches were temporarily downsampled to 256$\times$256 pixels and fed into the DeepLabv3 model, which performed pixel-wise classification to distinguish between tumor and non-tumor regions. Subsequently, only the patches contained more than 50\% tumor area were retained and stored as 1024$\times$1024 patches at a resolution of 0.549~µm per pixel.

\subsubsection{Histological Feature Extraction}
To alleviate the input resolution constraints of pretrained pathological feature extractors, we follow~\cite{cui2025prediction} to introduce the hierarchical UNI (hi-UNI) for multiple pathological FoVs feature extraction. Specifically, the original 1024$\times$1024 tumor patch was downsampled to 224$\times$224 at 2.509 microns per pixel (mpp) to represent macro-level morphological pattern; a 512$\times$512 region which center cropped from original 1024$\times$1024 patch was further downsampled to 224$\times$224 at 1.255 mpp to capture macroscopic histological feature and tissue-level representation. Finally, these two FoVs patch features were fed into the pathological foundation model UNI~\cite{chen2024uni} independently, for capturing global region patterns (macro-organization feature, $f_{\text{macro}}$) and local tissue details (meso-pheotype feature, $f_{\text{meso}}$) simultaneously. 

\subsubsection{Micro-Level Geometric Feature Extraction}
\label{micro-geometry-feature}
To capture micro cellular-level geometric structure within the TME, we leveraged Hover-Net~\cite{graham2019hover} to segment the nuclei in pathological images and classify them into five crucial categories in clinical practice: tumor, inflammatory, stroma, necrosis (dead) and epithelial (normal). In this work, we focus on tumor, inflammatory, stromal and epithelial nuclei, which represent the major and functionally relevant cellular components in the TME for liver cancer~\cite{dong2024spatial,yin2024single}. To this end, we extracted three types of handcrafted histological features $\mathbb{X}_{i}$ for each $i$-th nucleus: morphological features indicating cell’s shape and contour, texture features reflecting local pixel patterns via gray-level co-occurrence matrices (GLCM), and topological features characterizing intercellular relationships. Consequently, we design the micro-geometry feature to represent the micro cellular-level interactions via a geometric structure which can be regarded as the deepest FoV in the pyramid gigapixel WSI. We construct this geometric framework by employing a $k$-nearest neighbors ($k$-NN, $k=8$) algorithm to define edge connectivity, connecting each nucleus to its eight nearest neighbors within a 100-pixel (54.9 µm) distance threshold:
\begin{equation}
\mathrm{E} = \left\{ (\mathrm{V}_i, \mathrm{V}_j) \;\middle|\; \mathrm{V}_j \in \mathrm{kNN}(\mathrm{V}_i),\; \mathcal{D}(\mathrm{V}_i, \mathrm{V}_j) < \mathrm{T} \right\}
\end{equation}
where $\mathrm{V}_i$ indicate the nodes (nuclei) in the graph, and $\mathrm{kNN}(\cdot)$ denotes the set of $k$ nearest neighbors of node $\mathrm{V}_i$. $\mathcal{D}(\mathrm{V}_i, \mathrm{V}_j)$ indicates the Euclidean distance between node $\mathrm{V}_i$ and $\mathrm{V}_j$, and $\mathrm{T}$ is the threshold for edge length, set to 100 pixels (54.9 $\mu$m) in our study. Then, we can construct the binary adjacency matrix $\mathbb{A} \in \{0,1\}^{n \times n}$ based on $k$-NN:
\begin{equation}
\mathbb{A}_{ij} = \begin{cases}
1, & \text{if } \mathrm{V}_j \in \mathrm{kNN}(\mathrm{V}_i) \land \mathcal{D}(\mathrm{V}_i,\mathrm{V}_j) < \mathrm{T} \\
0, & \text{otherwise}
\end{cases}
\end{equation}
so that we can build a graph structure via the nucleus and the connectivity across these nucleus, which can be formulated as $\mathbb{G} = \left( \mathrm{V}, \mathrm{E}, \mathbb{X} \right)$, then we follow~\cite{kipf2016semi,zheng2022graph} to implemented a GCN layer to handle the graph structure and further transform it into a micro-geometry feature representation $f_{\textbf{g}}$.

\subsection{Hierarchical FoVs Alignment Module}
As aforementioned, pathological WSIs exhibit a hierarchical pyramid structure of visual features across varying resolutions: the features in lower FoV (e.g., 10$\times$) characterize macro organization in tissue (the extent of tumor-immune localization in describing tumor-infiltrating
versus tumor-distal lymphocytes), while high FoV features (e.g., 20$\times$) encompass the bounding box of cells and other tissue-level morphological features~\cite{chen2022scaling}. Therefore, we introduce a Hierarchical FoVs Alignment module to capture the hierarchical relationships and the crucial dependencies across image resolutions of WSIs. Specifically, we leverage a combination of MLP-Mixer~\cite{tolstikhin2021mlp} and Gated Attention Pooling Network~\cite{ilse2018attention}, which can enhance information communication and modeling of hierarchical structures as well as produce a contribution-weighted output for multi-FoVs feature aggregation. The detailed operation of our HFA module can be formulated as:
\begin{equation}
    \begin{aligned}
        & f_{\text{FoV}} = \mathrm{HFA} \left( f_{\text{c}} \right),
        f_{\text{c}} = \mathrm{Concat}\left( f_{\text{macro}}, f_{\text{meso}}\right) \\
        & \mathrm{HFA} \left( f_{\text{c}} \right) 
        = \mathrm{Linear} \Big( \mathrm{GAP} \big( \mathrm{MLP}_{\text{Mixer}} \left( f_{\text{c}} \right) \big)\Big)
    \end{aligned}
\end{equation}
where $\mathrm{MLP}_{\text{Mixer}}(\cdot)$ and $\mathrm{GAP}(\cdot)$ indicate the introduced MLP-Mixer and Gated Attention Pooling Network, respectively. The $\mathrm{MLP}_{\text{Mixer}}(\cdot)$ comprises a token-mixing MLP and a channel-mixing MLP, each implemented with two fully connected layers and a GELU activation function. The former enables cross-scale interaction between hierarchical features, while the latter facilitates intra-scale feature aggregation. This design promotes effective information communication across FoVs, further enhance the modeling for both intra-scale and inter-scale representations. The $\mathrm{MLP}_{\text{Mixer}}(\cdot)$ can be formulated as:
\begin{equation}
    \begin{aligned}
        & \mathrm{Z}_1 = f_{\text{c}}^{\top} 
        + \mathrm{Linear} \big( \Theta \left( f_{\text{c}}^{\top} \right) \cdot \mathrm{W}_{1} \big) \cdot \mathrm{W}_{2} \\
        & \mathrm{Z} = \mathrm{Z}_1^{\top}  
        + \mathrm{Linear} \big( \Theta \left( \mathrm{Z}_1^{\top} \right) \cdot \mathrm{W}_{3} \big) \cdot \mathrm{W}_{4}
    \end{aligned}
\end{equation}
where $\Theta(\cdot)$ denotes the combination of Layer Normalization and GELU, $\mathrm{W}_{1}, \mathrm{W}_{2}, \mathrm{W}_{3}$ and $\mathrm{W}_{4}$ are trainable weights of the fully connected linear layers. Then, the hidden state $\mathrm{Z}$ will be sent into a shared gated attention pooling network $\mathrm{GAP}(\cdot)$, which consists of two linear layers with ReLU and Sigmoid, the formulation of this procedure can be described as:
\begin{equation}
    f_{\text{FoV}} 
        = \alpha_{1} \cdot \Phi_{1}\left(\mathrm{Z}\right) 
        + \alpha_{2} \cdot \Phi_{2}\left(\mathrm{Z}\right), 
        \alpha_{i} = \mathrm{Sigmoid} \big( \Phi_{i}\left(\mathrm{Z}\right) \big)
\end{equation}
where $\Phi(\cdot)$ is a MLP layer with a architecture of Linear-ReLU-Linear-LayerNorm. We here compute the weighted importance scores assigned to each MLP branch via $\mathrm{Sigmoid}(\cdot)$. In this way, we can obtain an importance-weighted dynamic representation which reflecting the actual contributions for the two FoV features unlike normal fixed-scale fusion strategies.

\newcommand{\myboxsize}{1\textwidth}
\newcommand{\myarraystretch}{1.3}
\begin{table*}[t]
    \center
    \caption{The performance compared with 11 baselines on two datasets. ``M.'' indicates whether to use morphological features and ``G.'' indicates whether to use geometry features. The best and second best results are highlighted in {\color{red} red} and {\color{blue} blue}, respectively.}
	\renewcommand{\arraystretch}{\myarraystretch}
	\setlength\tabcolsep{6pt}
	\centering
	\resizebox{\myboxsize}{!}
    {
    \begin{tabular}{lccccccccccc}
        \toprule[1pt]
        {\multirow{2}{*}{Methods}} & \multicolumn{2}{c}{Modality} & \multicolumn{4}{c}{TCGA-Liver} & \multicolumn{4}{c}{DTH-ICC}
        \\ \cmidrule(lr){2-3} \cmidrule(lr){4-7} \cmidrule(lr){8-11}
        & M. & G. 
        & AUC $\uparrow$ & ACC $\uparrow$ & F1 $\uparrow$ & Pre. $\uparrow$ 
        & AUC $\uparrow$ & ACC $\uparrow$ & F1 $\uparrow$ & Pre.$\uparrow$ \\ 
        \midrule
        ABMIL~\cite{ilse2018attention}
        & $\checkmark$ & 
        & 98.2 $\pm$ 1.2 & 95.9 $\pm$ 0.7 & 87.0 $\pm$ 2.0 & 86.9 $\pm$ 4.7 & 85.2 $\pm$ 0.8 & 69.0 $\pm$ 0.3 & 68.5 $\pm$ 1.5 & 69.0 $\pm$ 1.3 \\
        DSMIL~\cite{li2021dual}
        & $\checkmark$ &
        & 97.9 $\pm$ 1.9 & 96.1 $\pm$ 1.0 & 86.8 $\pm$ 3.6 & 88.9 $\pm$ 6.6 & 84.9 $\pm$ 0.7 & 67.7 $\pm$ 1.7 & 67.3 $\pm$ 0.7 & 67.1 $\pm$ 1.3 \\
        CLAM-SB~\cite{lu2021data} 
        & $\checkmark$ &
        & 98.0 $\pm$ 0.6 & 94.0 $\pm$ 1.1 & 81.7 $\pm$ 2.6 & 77.7 $\pm$ 3.2 & 83.9 $\pm$ 1.8 & 66.5 $\pm$ 3.8 & 67.6 $\pm$ 2.4 & 68.2 $\pm$ 2.3 \\
        CLAM-MB~\cite{lu2021data} 
        & $\checkmark$ &
        & 98.3 $\pm$ 1.3 & 95.1 $\pm$ 1.0 & 82.0 $\pm$ 1.7 & 88.1 $\pm$ 2.5 & 85.9 $\pm$ 0.8 & 68.4 $\pm$ 3.8 & 68.9 $\pm$ 4.4 & 69.7 $\pm$ 3.5 \\
        TransMIL~\cite{shao2021transmil}
        & $\checkmark$ &
        & 98.1 $\pm$ 1.4 & 95.1 $\pm$ 2.1 & 86.4 $\pm$ 3.9 & 83.2 $\pm$ 5.1 & 83.4 $\pm$ 2.5 & 63.4 $\pm$ 4.9 & 64.1 $\pm$ 4.3 & 67.0 $\pm$ 2.3 \\
        ACMIL~\cite{zhang2024attention}
        & $\checkmark$ &
        & 98.5 $\pm$ 1.6 & 96.3 $\pm$ 0.9 & 87.5 $\pm$ 1.9 & 90.4 $\pm$ 1.3 & 86.2 $\pm$ 1.2 & 69.5 $\pm$ 1.2 & 69.1 $\pm$ 1.7 & 70.0 $\pm$ 1.7 \\
        IBMIL~\cite{lin2023interventional}
        & $\checkmark$ &
        & 98.1 $\pm$ 1.5 & 95.5 $\pm$ 1.2 & 85.1 $\pm$ 1.8 & 88.4 $\pm$ 3.3 & 84.1 $\pm$ 1.8 & 66.5 $\pm$ 1.2 & 67.2 $\pm$ 1.9 & 67.4 $\pm$ 1.0 \\
        MHIM-MIL~\cite{tang2023multiple}
        & $\checkmark$ &
        & 98.7 $\pm$ 0.8 & 96.4 $\pm$ 0.5 & 89.0 $\pm$ 2.6 & 86.8 $\pm$ 2.7 & 86.4 $\pm$ 1.5 & 69.1 $\pm$ 2.2 & 70.0 $\pm$ 0.9 & 69.9 $\pm$ 1.1 \\
        \midrule 
        Patch-GCN~\cite{chen2021whole}
        & & $\checkmark$
        & 96.2 $\pm$ 3.2 & 94.1 $\pm$ 1.9 & 78.4 $\pm$ 3.0 & 85.2 $\pm$ 3.1 & 74.7 $\pm$ 1.5 & 59.5 $\pm$ 2.5 & 58.2 $\pm$ 2.2 & 60.7 $\pm$ 1.2 \\
        GTMIL~\cite{zheng2022graph}
        & & $\checkmark$
        & 97.9 $\pm$ 1.5 & 96.4 $\pm$ 1.3& 85.4 $\pm$ 2.0& \color{blue}93.1 $\pm$ 2.5 & 82.6 $\pm$ 1.4 & 61.3 $\pm$ 2.1 & 61.5 $\pm$ 1.7 & 61.4 $\pm$ 0.9 \\
        NPKC-MIL~\cite{wang2024nuclei}
        & $\checkmark$ & $\checkmark$
        & \color{blue}99.1 $\pm$ 1.1 & \color{blue}96.9 $\pm$ 0.9 & \color{blue}91.8 $\pm$ 2.3 & 91.2 $\pm$ 3.7 & \color{blue}86.8 $\pm$ 2.5 & \color{blue}70.2 $\pm$ 2.2 & \color{blue}70.6 $\pm$ 2.3 & \color{blue}71.8 $\pm$ 1.9 \\
        \midrule
        \textbf{ARGUS (Ours)} 
        & $\checkmark$ & $\checkmark$
        & \color{red}99.5 $\pm$ 0.3 & \color{red}98.1 $\pm$ 0.7 & \color{red}93.5 $\pm$ 2.3 & \color{red}93.6 $\pm$ 5.3 & \color{red}88.4 $\pm$ 1.3 & \color{red}74.0 $\pm$ 0.9 & \color{red}73.7 $\pm$ 0.6 & \color{red}74.6 $\pm$ 0.4 \\
        \bottomrule[1pt]
    \end{tabular}
    }\label{tab:comparison}
\end{table*}

\subsection{Geometry Prior Guided Fusion Module}
To learn the fine-grained contextual relationship among all the nucleus within the TME, we propose a novel geometry prior guided attention operation, termed GPGF, aiming to treat the geometric structure as a geometric knowledge prior to guide the feature integration. The precomputed micro-geometry feature can be considered the deepest fine-grained cell/nuclei-level FoV of input WSI, thereby we can construct a geometry-aware overall FoVs feature with the combination of the macro-meso FoVs fusion feature $f_{\text{FoV}}$ and micro-geometry feature $f_{\text{g}}$. Then, for modeling the holistic interactions between geometry-feature $f_{\text{g}}$ and morphological feature $f_{\text{FoV}}$, we perform this feature integration operation in a "intra-modality with inter-modality" manner as follows:
\begin{equation}
    \begin{aligned}
        f_{\text{FoV}}^{\text{self}} 
        &= \Gamma \left( f_{\text{FoV}}, f_{\text{FoV}}, f_{\text{FoV}} \right), 
        f_{\text{FoV}}^{\text{cross}} 
        = \Gamma \left( f_{g \rightarrow \text{FoV}}, f_{\text{FoV}}, f_{\text{FoV}} \right), \\
        f_{g}^{\text{self}} 
        &= \Gamma \left( f_{\text{g}}, f_{\text{g}}, f_{\text{g}} \right),
        f_{\text{g}}^{\text{cross}} 
        = \Gamma \left( f_{FoV \rightarrow \text{g}}, f_{\text{g}}, f_{\text{g}} \right),
    \end{aligned}
\end{equation}
where $\Gamma(\cdot)$ indicates the Multi-Head Attention (MHA) mechanism~\cite{vaswani2017attention}, we then perform a gating strategy to compute the balanced representation dynamically as following:
\begin{equation}
    \begin{aligned}
        f^{\prime}_{\text{FoV}} 
        & = \alpha_{\text{FoV}} \cdot f_{\text{FoV}}^{\text{self}}
        + (1 - \alpha_{\text{FoV}}) \cdot f_{\text{FoV}}^{\text{cross}} \\
        f^{\prime}_{\text{g}} 
        & = \alpha_{\text{g}} \cdot f_{\text{g}}^{\text{self}}
        + (1 - \alpha_{\text{g}}) \cdot f_{\text{g}}^{\text{cross}}
    \end{aligned}
\end{equation}
we further combine the enhanced representations $f^{\prime}_{\text{FoV}}$ and $f^{\prime}_{\text{g}}$ then send it into a Transformer Layer~\cite{dosovitskiy2020image} for modeling the long-range dependencies within the final representation:
\begin{equation}
    f_{\text{out}} = 
    \mathrm{TransLayer} \big( 
    \mathrm{Concat} \left( f^{\prime}_{\text{FoV}}, f^{\prime}_{\text{g}} \right) 
    \big)
\end{equation}
Lastly, a fully-connected layer is employed to produce the final representation $f_{\text{out}}$ for histological subtyping of liver cancer.

\section{Experiments and Results}
\subsection{Datasets}
To rigorously evaluate the efficacy, robustness, and clinical applicability of our model, we curated a diverse set of two WSI datasets on liver cancer subtyping, encompassing both publicly and in-house collections, including a dataset from TCGA Data portal\footnote{TCGA: \url{https://portal.gdc.cancer.gov}}: \textbf{TCGA-Liver}, which is curated for liver caner subtyping (HCC vs. ICC) and composed of 413 WSIs from the TCGA-LIHC project (Hepatocellular Carcinoma, 379 WSIs from 365 patients) and the TCGA-CHOL (Intrahepatic Cholangiocarcinoma, 34 WSIs from 34 patients) project.
To validate the generalizability of ARGUS, we also incorporated an in-house cohort: \textbf{DTH-ICC}, a histology WSI dataset for ICC fine-grained subtyping, which comprises 789 WSIs collected from Department of Pathology, Nanjing Drum Tower Hospital, Affiliated Hospital of Medical School, Nanjing University, Nanjing, China, which includes fine-duct (289 WSIs from 67 patients), small-duct (241 WSIs from 78 patients) and large-duct (259 WSIs from 115 patients) three subtypes.

\subsection{Implementation Details}
\subsubsection{Training settings} 
We select a diverse set of baselines, including those focused on visual feature based MILs and geometric feature based models. The methods for comparison include: ABMIL~\cite{ilse2018attention}, DSMIL~\cite{li2021dual}, CLAM-SB~\cite{lu2021data}, CLAM-MB~\cite{lu2021data},  TransMIL~\cite{shao2021transmil}, ACMIL~\cite{zhang2024attention}, IBMIL~\cite{lin2023interventional}, MHIM-MIL~\cite{tang2023multiple}, Patch-GCN~\cite{chen2021whole}, GTMIL~\cite{zheng2022graph}, and NPKC-MIL~\cite{wang2024nuclei}. To evaluate ARGUS, we follow standard practice to conduct experiments using 5-fold Monte-Carlo cross-validation to alleviate the batch effect. The accuracy (ACC), Area Under the Curve (AUC), F1-Score (F1), and Precision (Pre.) four metrics were employed to measure the diagnosis ability of the models. 

\subsubsection{Hyper-parameters}
The ARGUS model was built using the PyTorch framework and trained on a GeForce RTX 4090 GPU workstation. During the training process of the ARGUS model, cross-entropy is used as the loss function, the batch size was set to 10, and the AdamW optimizer with a weight decay of 1e–3 and a learning rate of 2e–5 was employed.

\begin{table*}[t]
	\centering
	\caption{Quantitative results for ablation study on two datasets. We \textbf{bold} the highest performance.}\label{tab:ablation}
	\renewcommand{\arraystretch}{\myarraystretch}
	\setlength\tabcolsep{6pt}
	\resizebox{\myboxsize}{!}
	{\begin{tabular}{cccccccccc}
    \toprule[1pt]
    \multirow{2}{*}{Model} & \multicolumn{3}{c}{{Designs in our model}} & \multicolumn{3}{c}{TCGA-Liver} & \multicolumn{3}{c}{DTH-ICC} \\
    \cmidrule(lr){2-4} \cmidrule(lr){5-7} \cmidrule(lr){8-10}
    & HFA & Geometry Feature & GPGF   
    & AUC $\uparrow$ & ACC $\uparrow$ & F1 $\uparrow$ 
    & AUC $\uparrow$ & ACC $\uparrow$ & F1 $\uparrow$ \\ 
    \midrule
    A & & & & 98.1 $\pm$ 0.8 & 94.9 $\pm$ 1.1 & 78.6 $\pm$ 5.3 & 
    85.0 $\pm$ 0.9 & 66.8 $\pm$ 3.5 & 66.9 $\pm$ 3.2 \\
    B & $\checkmark$ & & & 98.3 $\pm$ 0.6 & 95.1 $\pm$ 0.5 & 86.1 $\pm$ 1.6 & 86.9 $\pm$ 1.5 & 70.9 $\pm$ 3.0 & 70.6 $\pm$ 4.1 \\
    C & & $\checkmark$ & & 98.3 $\pm$ 0.9 & 95.1 $\pm$ 0.8 & 86.6 $\pm$ 1.2 & 86.6 $\pm$ 1.5 & 69.0 $\pm$ 4.3 & 68.2 $\pm$ 4.5 \\
    D & $\checkmark$ & $\checkmark$ & & 98.7 $\pm$ 0.3 & 96.3 $\pm$ 0.1 & 89.0 $\pm$ 0.7 & 87.1 $\pm$ 2.4 & 70.9 $\pm$ 3.5 & 70.6 $\pm$ 3.2 \\
    E & & $\checkmark$ & $\checkmark$ & 99.1 $\pm$ 0.4 & 96.4 $\pm$ 0.6 & 90.2 $\pm$ 0.9& 87.3 $\pm$ 1.2 & 71.1 $\pm$ 1.5 & 71.2 $\pm$ 1.2 \\
    F & $\checkmark$ & $\checkmark$ & $\checkmark$ & \textbf{99.5 $\pm$ 0.3} & \textbf{98.1 $\pm$ 0.7} & \textbf{93.5 $\pm$ 2.3} & 
    \textbf{88.4 $\pm$ 1.3} & \textbf{74.0 $\pm$ 0.9} & \textbf{73.7 $\pm$ 0.6} \\
	\toprule[1pt]
	\end{tabular}}
\end{table*}

\subsection{Comparison with State-of-the-art Methods}
To demonstrate the advantages of our proposed ARGUS, we conducted extensive experiments compared with 11 cutting-edge baselines using identical settings on both public and in-house cohorts. As shown in Tab.~\ref{tab:comparison}, ARGUS achieves superior performance for histological subtyping on both two datasets. Against TransMIL~\cite{shao2021transmil}, the current SOTA MIL method, our model achieves the performance increases of 1.43\% on AUC, 3.15\% on Accuracy, 7.37\% on F1-Score, and 12.5\% on Precision on TCGA-Liver dataset, respectively. This suggests that histological subtyping should focus on the hierarchical structure of phenotypes in the TME, rather than single-level low-resolution image features. Notably, most traditional MILs steadily outperform geometry-only methods, highlighting the crucial contribution of integrating information from both morphological and geometric features. Additionally, our ARGUS also outperforms NPKC-MIL~\cite{wang2024nuclei}, a morphological-geometric multimodal counterpart, further emphasizing the importance of hierarchical pyramid structure and advanced geometric representations in the TME. Finally, our model consistently outperforms other SOTA histological subtyping methods by a large margin on both two datasets.

\subsection{Ablation Study}
To systematically evaluate the effectiveness of each modules in our proposed ARGUS framework, we conducted a series of ablation experiments, as summarized in Tab.~\ref{tab:ablation}. We started with a basic model (Model A) based on the simple weakly-supervised MIL baseline using histopathological features.


\smallskip
\noindent{\bf Hierarchical FoVs Alignment (HFA) Strategy.}  
To assess the contribution of the Hierarchical FoVs Alignment (HFA) strategy, we first compared Model B and Model A. The inclusion of HFA improved AUC by 0.20\% and 2.23\% on the TCGA-Liver and DTH-ICC datasets, respectively. This suggests that incorporating hierarchical FoVs features allows the model to better capture complementary histological cues at different FoVs. 
To further verify the robustness of HFA, we compared Model D (with both HFA and geometry features) against Model C (with only geometry features). We observed additional AUC gains of 0.41\% on TCGA-Liver and 0.58\% on DTH-ICC, confirming that the benefit of HFA persists when micro-Level geometry representations are incorporated. Finally, we compared the full model (Model F), which integrates all components including HFA, with Model E (without HFA). Model F achieved AUC improvements of 0.40\% and 1.26\% on the two cohorts, respectively. These consistent improvements across multiple settings validate the effectiveness of HFA.

\smallskip
\noindent{\bf Micro-Level Geometry Feature Usage.} 
To assess the effectiveness of our designed micro-geometry feature which produced by building a $k$-NN graph using hand-crafted pathological features, we obtain the Model C by introducing the micro-geometry feature into Model A. Notably, in DTH-ICC dataset, Model C exhibited a significant improvement of 1.8\% in AUC performance over Model A. This outcome demonstrates the significance of integrating micro-level geometry feature, as it proves to be indispensable in enhancing the overall histological subtyping performance of ARGUS.


\smallskip
\noindent{\bf Geometry Prior Guided Attention (GPGF) Strategy.}  
We proceeded to augment Model C by incorporating the Geometry Prior Guided Attention (GPGF) strategy to create Model E, followed by a comparative evaluation between the resulting Model E and Model C (w/o GPGF). The experiment result highlighted the critical contribution of the GPGF module. Removing the GPGF operation (Model C) resulted in significant degradation of performance, which notably affected the AUC on two datasets. Therefore, the geometry prior guided attention strategy effectively fuses hierarchical FoVs and micro-level geometry features, substantially enhancing the overall representational capacity of ARGUS.


\begin{figure*}[t]
    \centering    
    \centerline{\includegraphics[width=1\textwidth]{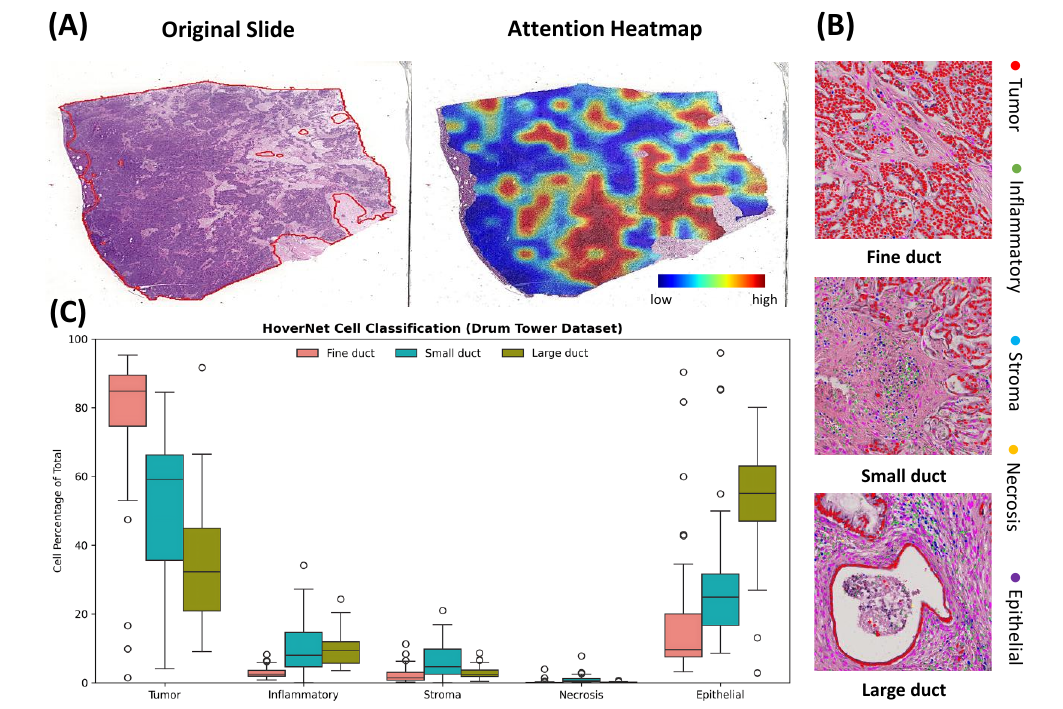}}
    \caption{Model visualization and interpretability analysis of the proposed ARGUS on DTH-ICC dataset. (a) the input WSI, associated corresponding attention heatmap for histological subtyping. (b) Representative high-attention patches from three ICC subtypes, overlaid with corresponding cell-type annotations. (c) Quantitative analysis of cell types in the top 10\% high-attention patches.}
    \label{visual_inter}
\end{figure*}

\subsection{Visualization and Interpretability Analysis}
We generated the attention heatmaps assembling the tiles extracted from tumor regions within each WSI and assigning the corresponding attention scores to create a mosaic mask. The tile-level attention scores were directly used to construct the heatmaps. To mitigate the artifacts introduced by tile boundaries, Gaussian filtering was applied for smooth visualization. Fig.~\ref{visual_inter}(A) shows the resulting heatmaps alongside their corresponding original WSIs. The darker red regions in the heatmaps indicate areas with higher attention scores, usually considered by the model to be of greater diagnostic value which typically align well with the clinical characteristics of different subtypes. In contrast, regions with lower attention values are typically concentrated within compact tumor nests, where cellular morphology tends to be uniform and lacks distinctive subtype-specific features which contributed limited discriminative information for subtyping task.

We further investigated the cellular distribution by performing nuclei segmentation and classification to the top 10\% of high-attention patches from three ICC subtypes (Fig.~\ref{visual_inter}(B)) on DTH-ICC cohort. Color-coded overlays reveal the spatial distribution of five nuclei types within these regions. We also quantified the distribution of all cell types across different subtypes, revealing subtype-specific distribution patterns. In Fine duct cases, cells predominantly distribute in tumor cell-enriched regions. In contrast, Large duct cases tend to focus on areas with dense epithelial cell populations, with Small duct cases presenting intermediate patterns in the distribution of these cell types. The related boxplot (Fig.~\ref{visual_inter}(C)) further confirms statistically significant differences among subtypes, underscoring the heterogeneity in cellular organization.

These results confirm that integrating hierarchical FoVs and geometric features improves not only diagnostic performance but interpretability on biological relevant subtype distinctions.

\section{Conclusion}
\label{sec:conclusion}
In this paper, we propose \textbf{A} hie\textbf{R}archical \textbf{G}eometry-g\textbf{U}ided tran\textbf{S}former (ARGUS) to comprehensively model the macro–meso–micro hierarchical interactions within histopathological whole slide images (WSIs) for the histological subtyping of primary liver malignancies. We introduce a novel Hierarchical FoVs Alignment (HFA) module that integrates macro- and meso-scale pathological features through a contribution-weighted dynamic fusion strategy. Furthermore, by leveraging a geometry prior guided attention mechanism, ARGUS effectively fuses hierarchical FoVs histological information and micro-level geometric representation cues to capture complementary morphological and cellular-level patterns within TME. Extensive experiments conducted on both public and private cohorts demonstrate the effectiveness of our proposed ARGUS framework.

{\small
\bibliographystyle{IEEEtranS}
\bibliography{ref}

\begin{thebibliography}{10}
\providecommand{\url}[1]{#1}
\csname url@samestyle\endcsname
\providecommand{\newblock}{\relax}
\providecommand{\bibinfo}[2]{#2}
\providecommand{\BIBentrySTDinterwordspacing}{\spaceskip=0pt\relax}
\providecommand{\BIBentryALTinterwordstretchfactor}{4}
\providecommand{\BIBentryALTinterwordspacing}{\spaceskip=\fontdimen2\font plus
\BIBentryALTinterwordstretchfactor\fontdimen3\font minus \fontdimen4\font\relax}
\providecommand{\BIBforeignlanguage}[2]{{%
\expandafter\ifx\csname l@#1\endcsname\relax
\typeout{** WARNING: IEEEtranS.bst: No hyphenation pattern has been}%
\typeout{** loaded for the language `#1'. Using the pattern for}%
\typeout{** the default language instead.}%
\else
\language=\csname l@#1\endcsname
\fi
#2}}
\providecommand{\BIBdecl}{\relax}
\BIBdecl

\bibitem{akinyemiju2017burden}
T.~Akinyemiju, S.~Abera, M.~Ahmed, N.~Alam, M.~A. Alemayohu, C.~Allen, R.~Al-Raddadi, N.~Alvis-Guzman, Y.~Amoako, A.~Artaman \emph{et~al.}, ``The burden of primary liver cancer and underlying etiologies from 1990 to 2015 at the global, regional, and national level: results from the global burden of disease study 2015,'' \emph{JAMA oncology}, vol.~3, no.~12, pp. 1683--1691, 2017.

\bibitem{bray2024global}
F.~Bray, M.~Laversanne, H.~Sung, J.~Ferlay, R.~L. Siegel, I.~Soerjomataram, and A.~Jemal, ``Global cancer statistics 2022: Globocan estimates of incidence and mortality worldwide for 36 cancers in 185 countries,'' \emph{CA: a cancer journal for clinicians}, vol.~74, no.~3, pp. 229--263, 2024.

\bibitem{cai2024seqfrt}
C.~Cai, J.~Li, M.~Liu, Y.~Jiao, and J.~Xu, ``Seqfrt: Towards effective adaption of foundation model via sequence feature reconstruction in computational pathology,'' in \emph{2024 IEEE International Conference on Bioinformatics and Biomedicine (BIBM)}.\hskip 1em plus 0.5em minus 0.4em\relax IEEE, 2024, pp. 1808--1815.

\bibitem{calderaro2023deep}
J.~Calderaro, N.~Ghaffari~Laleh, Q.~Zeng, P.~Maille, L.~Favre, A.~Pujals, C.~Klein, C.~Bazille, L.~R. Heij, A.~Uguen \emph{et~al.}, ``Deep learning-based phenotyping reclassifies combined hepatocellular-cholangiocarcinoma,'' \emph{Nature communications}, vol.~14, no.~1, p. 8290, 2023.

\bibitem{chen2017rethinking}
L.-C. Chen, G.~Papandreou, F.~Schroff, and H.~Adam, ``Rethinking atrous convolution for semantic image segmentation,'' \emph{arXiv preprint arXiv:1706.05587}, 2017.

\bibitem{chen2022scaling}
R.~J. Chen, C.~Chen, Y.~Li, T.~Y. Chen, A.~D. Trister, R.~G. Krishnan, and F.~Mahmood, ``Scaling vision transformers to gigapixel images via hierarchical self-supervised learning,'' in \emph{Proceedings of the IEEE/CVF Conference on Computer Vision and Pattern Recognition}, 2022, pp. 16\,144--16\,155.

\bibitem{chen2024uni}
R.~J. Chen, T.~Ding, M.~Y. Lu, D.~F. Williamson, G.~Jaume, B.~Chen, A.~Zhang, D.~Shao, A.~H. Song \emph{et~al.}, ``Towards a general-purpose foundation model for computational pathology,'' \emph{Nature Medicine}, 2024.

\bibitem{chen2021whole}
R.~J. Chen, M.~Y. Lu, M.~Shaban, C.~Chen, T.~Y. Chen, D.~F. Williamson, and F.~Mahmood, ``Whole slide images are 2d point clouds: Context-aware survival prediction using patch-based graph convolutional networks,'' in \emph{Medical Image Computing and Computer Assisted Intervention--MICCAI 2021: 24th International Conference, Strasbourg, France, September 27--October 1, 2021, Proceedings, Part VIII 24}.\hskip 1em plus 0.5em minus 0.4em\relax Springer, 2021, pp. 339--349.

\bibitem{cui2025prediction}
H.~Cui, Q.~Guo, J.~Xu, X.~Wu, C.~Cai, Y.~Jiao, W.~Ming, H.~Wen, and X.~Wang, ``Prediction of molecular subtypes for endometrial cancer based on hierarchical foundation model,'' \emph{Bioinformatics}, p. btaf059, 2025.

\bibitem{dong2024spatial}
Z.-R. Dong, M.-Y. Zhang, L.-X. Qu, J.~Zou, Y.-H. Yang, Y.-L. Ma, C.-C. Yang, X.-L. Cao, L.-Y. Wang, X.-L. Zhang \emph{et~al.}, ``Spatial resolved transcriptomics reveals distinct cross-talk between cancer cells and tumor-associated macrophages in intrahepatic cholangiocarcinoma,'' \emph{Biomarker Research}, vol.~12, no.~1, p. 100, 2024.

\bibitem{dosovitskiy2020image}
A.~Dosovitskiy, L.~Beyer, A.~Kolesnikov, D.~Weissenborn, X.~Zhai, T.~Unterthiner, M.~Dehghani, M.~Minderer, G.~Heigold, S.~Gelly \emph{et~al.}, ``An image is worth 16x16 words: Transformers for image recognition at scale,'' \emph{arXiv preprint arXiv:2010.11929}, 2020.

\bibitem{european2023easl}
E.~A. for the Study~of The~Liver \emph{et~al.}, ``Easl-ilca clinical practice guidelines on the management of intrahepatic cholangiocarcinoma,'' \emph{Journal of hepatology}, vol.~79, no.~1, pp. 181--208, 2023.

\bibitem{graham2019hover}
S.~Graham, Q.~D. Vu, S.~E.~A. Raza, A.~Azam, Y.~W. Tsang, J.~T. Kwak, and N.~Rajpoot, ``Hover-net: Simultaneous segmentation and classification of nuclei in multi-tissue histology images,'' \emph{Medical image analysis}, vol.~58, p. 101563, 2019.

\bibitem{hu2019comparative}
J.~Hu, H.~Zhou, W.~Liu, J.~Zhang, H.~Hu, and J.~Liu, ``A comparative study of intrahepatic cholangiocarcinoma and hepatocellular carcinoma with reference to clinical features and prognosis,'' \emph{Zhonghua gan Zang Bing za zhi= Zhonghua Ganzangbing Zazhi= Chinese Journal of Hepatology}, vol.~27, no.~7, pp. 511--515, 2019.

\bibitem{ilse2018attention}
M.~Ilse, J.~Tomczak, and M.~Welling, ``Attention-based deep multiple instance learning,'' in \emph{International conference on machine learning}.\hskip 1em plus 0.5em minus 0.4em\relax PMLR, 2018, pp. 2127--2136.

\bibitem{kipf2016semi}
T.~Kipf, ``Semi-supervised classification with graph convolutional networks,'' \emph{arXiv preprint arXiv:1609.02907}, 2016.

\bibitem{li2021dual}
B.~Li, Y.~Li, and K.~W. Eliceiri, ``Dual-stream multiple instance learning network for whole slide image classification with self-supervised contrastive learning,'' in \emph{Proceedings of the IEEE/CVF conference on computer vision and pattern recognition}, 2021, pp. 14\,318--14\,328.

\bibitem{lin2023interventional}
T.~Lin, Z.~Yu, H.~Hu, Y.~Xu, and C.-W. Chen, ``Interventional bag multi-instance learning on whole-slide pathological images,'' in \emph{Proceedings of the IEEE/CVF Conference on Computer Vision and Pattern Recognition}, 2023, pp. 19\,830--19\,839.

\bibitem{liu2025murrenet}
M.~Liu, C.~Cai, J.~Li, P.~Xu, J.~Li, J.~Ma, and J.~Xu, ``Murrenet: Modeling holistic multimodal interactions between histopathology and genomic profiles for survival prediction,'' \emph{arXiv preprint arXiv:2507.04891}, 2025.

\bibitem{liu2023mgct}
M.~Liu, Y.~Liu, H.~Cui, C.~Li, and J.~Ma, ``Mgct: Mutual-guided cross-modality transformer for survival outcome prediction using integrative histopathology-genomic features,'' in \emph{2023 IEEE International Conference on Bioinformatics and Biomedicine (BIBM)}.\hskip 1em plus 0.5em minus 0.4em\relax IEEE, 2023, pp. 1306--1312.

\bibitem{liu2024exploiting}
M.~Liu, Y.~Liu, P.~Xu, H.~Cui, J.~Ke, and J.~Ma, ``Exploiting geometric features via hierarchical graph pyramid transformer for cancer diagnosis using histopathological images,'' \emph{IEEE Transactions on Medical Imaging}, 2024.

\bibitem{liu2024unleashing}
M.~Liu, Y.~Liu, P.~Xu, and J.~Ma, ``Unleashing the infinity power of geometry: A novel geometry-aware transformer (goat) for whole slide histopathology image analysis,'' in \emph{2024 IEEE International Symposium on Biomedical Imaging (ISBI)}.\hskip 1em plus 0.5em minus 0.4em\relax IEEE, 2024, pp. 1--5.

\bibitem{lu2021data}
M.~Y. Lu, D.~F. Williamson, T.~Y. Chen, R.~J. Chen, M.~Barbieri, and F.~Mahmood, ``Data-efficient and weakly supervised computational pathology on whole-slide images,'' \emph{Nature biomedical engineering}, vol.~5, no.~6, pp. 555--570, 2021.

\bibitem{paradis2023pathogenesis}
V.~Paradis and J.~Zucman-Rossi, ``Pathogenesis of primary liver carcinomas,'' \emph{Journal of Hepatology}, vol.~78, no.~2, pp. 448--449, 2023.

\bibitem{shao2021transmil}
Z.~Shao, H.~Bian, Y.~Chen, Y.~Wang, J.~Zhang, X.~Ji \emph{et~al.}, ``Transmil: Transformer based correlated multiple instance learning for whole slide image classification,'' \emph{Advances in neural information processing systems}, vol.~34, pp. 2136--2147, 2021.

\bibitem{song2025deep}
S.~Song, G.~Zhang, Z.~Yao, R.~Chen, K.~Liu, T.~Zhang, G.~Zeng, Z.~Wang, and R.~Liu, ``Deep learning based on intratumoral heterogeneity predicts histopathologic grade of hepatocellular carcinoma,'' \emph{BMC cancer}, vol.~25, no.~1, p. 497, 2025.

\bibitem{tang2023multiple}
W.~Tang, S.~Huang, X.~Zhang, F.~Zhou, Y.~Zhang, and B.~Liu, ``Multiple instance learning framework with masked hard instance mining for whole slide image classification,'' in \emph{Proceedings of the IEEE/CVF International Conference on Computer Vision}, 2023, pp. 4078--4087.

\bibitem{tolstikhin2021mlp}
I.~O. Tolstikhin, N.~Houlsby, A.~Kolesnikov, L.~Beyer, X.~Zhai, T.~Unterthiner, J.~Yung, A.~Steiner, D.~Keysers, J.~Uszkoreit \emph{et~al.}, ``Mlp-mixer: An all-mlp architecture for vision,'' \emph{Advances in neural information processing systems}, vol.~34, pp. 24\,261--24\,272, 2021.

\bibitem{vaswani2017attention}
A.~Vaswani, N.~Shazeer, N.~Parmar, J.~Uszkoreit, L.~Jones, A.~N. Gomez, {\L}.~Kaiser, and I.~Polosukhin, ``Attention is all you need,'' \emph{Advances in neural information processing systems}, vol.~30, 2017.

\bibitem{wang2024nuclei}
X.~Wang and W.~Yuan, ``Nuclei-level prior knowledge constrained multiple instance learning for breast histopathology whole slide image classification,'' \emph{Iscience}, vol.~27, no.~6, 2024.

\bibitem{yang2025explainable}
Z.~Yang, C.~Guo, J.~Li, Y.~Li, L.~Zhong, P.~Pu, T.~Shang, L.~Cong, Y.~Zhou, G.~Qiao \emph{et~al.}, ``An explainable multimodal artificial intelligence model integrating histopathological microenvironment and ehr phenotypes for germline genetic testing in breast cancer,'' \emph{Advanced Science}, p. e02833, 2025.

\bibitem{yin2024single}
Z.~Yin, Y.~Song, and L.~Wang, ``Single-cell rna sequencing reveals the landscape of the cellular ecosystem of primary hepatocellular carcinoma,'' \emph{Cancer Cell International}, vol.~24, no.~1, p. 379, 2024.

\bibitem{zhang2024attention}
Y.~Zhang, H.~Li, Y.~Sun, S.~Zheng, C.~Zhu, and L.~Yang, ``Attention-challenging multiple instance learning for whole slide image classification,'' in \emph{European Conference on Computer Vision}.\hskip 1em plus 0.5em minus 0.4em\relax Springer, 2024, pp. 125--143.

\bibitem{zheng2022graph}
Y.~Zheng, R.~H. Gindra, E.~J. Green, E.~J. Burks, M.~Betke, J.~E. Beane, and V.~B. Kolachalama, ``A graph-transformer for whole slide image classification,'' \emph{IEEE transactions on medical imaging}, vol.~41, no.~11, pp. 3003--3015, 2022.

\end{thebibliography}
}



\end{document}